%% file: egopose.tex
\documentclass[sigconf]{acmart}
\def\pp{Ego3DPose}

\usepackage{diagbox}

\setcopyright{acmlicensed}

\acmDOI{10.1145/3610548.3618147}

\copyrightyear{2023}
\acmYear{2023}
\setcopyright{acmlicensed}\acmConference[SA Conference Papers'23]{SIGGRAPH Asia 2023 Conference Papers}{December 12--15, 2023}{Sydney, NSW, Australia}
\acmBooktitle{SIGGRAPH Asia 2023 Conference Papers (SA Conference Papers'23), December 12--15, 2023, Sydney, NSW, Australia}
\acmPrice{15.00}
\acmISBN{979-8-4007-0315-7/23/12}

\usepackage{acmart-taps}

\begin{document}
\title{Ego3DPose: Capturing 3D Cues from Binocular Egocentric Views}

\author{Taeho Kang}
\affiliation{%
   \institution{Seoul National University}
   \city{Seoul}
   \country{South Korea}}
\email{taeho.kang@hcs.snu.ac.kr}

\author{Kyungjin Lee}
\affiliation{%
   \institution{Seoul National University}
   \city{Seoul}
   \country{South Korea}}
\email{jin11542@snu.ac.kr}

\author{Jinrui Zhang}
\affiliation{%
   \institution{Central South University}
   \city{Changsha}
   \country{China}}
\email{zhangjinruicsu@gmail.com}

\author{Youngki Lee}
\affiliation{%
   \institution{Seoul National University}
   \city{Seoul}
   \country{South Korea}}
\email{youngkilee@snu.ac.kr}

\begin{abstract}
We present \pp{}, a highly accurate binocular egocentric 3D pose reconstruction system. The binocular egocentric setup offers practicality and usefulness in various applications, however, it remains largely under-explored. It has been suffering from low pose estimation accuracy due to viewing distortion, severe self-occlusion, and limited field-of-view of the joints in egocentric 2D images. Here, we notice that two important 3D cues, stereo correspondences, and perspective, contained in the egocentric binocular input are neglected. Current methods heavily rely on 2D image features, implicitly learning 3D information, which introduces biases towards commonly observed motions and leads to low overall accuracy. We observe that they not only fail in challenging occlusion cases but also in estimating visible joint positions. To address these challenges, we propose two novel approaches. First, we design a two-path network architecture with a path that estimates pose per limb independently with its binocular heatmaps. Without full-body information provided, it alleviates bias toward trained full-body distribution. Second, we leverage the egocentric view of body limbs, which exhibits strong perspective variance (e.g., a significantly large-size hand when it is close to the camera). We propose a new perspective-aware representation using trigonometry, enabling the network to estimate the 3D orientation of limbs. Finally, we develop an end-to-end pose reconstruction network that synergizes both techniques. Our comprehensive evaluations demonstrate that \pp{} outperforms state-of-the-art models by a pose estimation error (i.e., MPJPE) reduction of 23.1\% in the UnrealEgo dataset. Our qualitative results highlight the superiority of our approach across a range of scenarios and challenges.

\end{abstract}
%
%


\begin{teaserfigure}
     \includegraphics[width=1.0\textwidth]
    {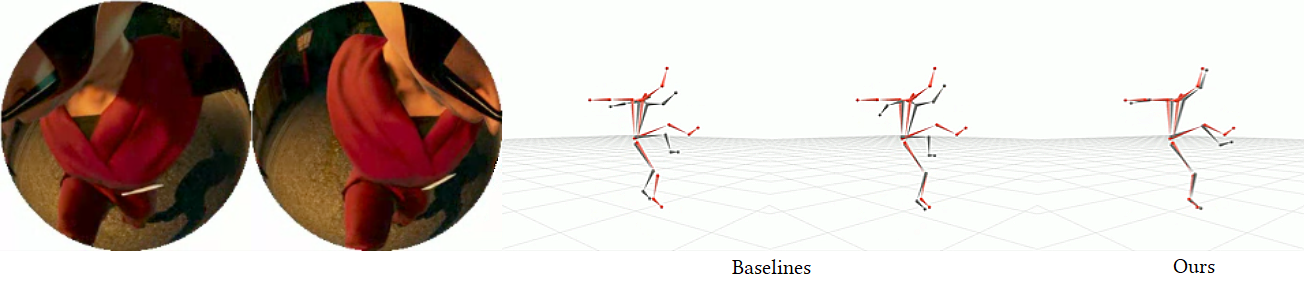}
\caption{The binocular egocentric input and a qualitative result of 3D pose estimation of previous methods and our system. The ground truth pose is in red, and the gray one corresponds to \cite{zhao2021egoglass}, \cite{hakada2022unrealego}, and our system from left to right.}
\label{fig:teaser}
\end{teaserfigure}

\maketitle
\input{01_introduction.tex}
\input{03_related_work.tex}
\input{04_method.tex}
\input{05_training.tex}
\input{06_evaluation.tex}
\input{07_conclusion.tex}

\bibliographystyle{ACM-Reference-Format}
\bibliography{egopose}
\input{appendix.tex}
\input{99_figure_only.tex}
\end{document}

%% file: 01_introduction.tex
\section{Introduction} 
\label{introduction}

Egocentric 3D pose reconstruction brings mobility to many promising applications such as avatar creation~\cite{ahuja2019mecap}, motion analysis for sports~\cite{hwang2017athelete,zecha2018sports} and health~\cite{biomac3d2023}, and action recognition~\cite{egoaction2022}. Among various egocentric setups~\cite{xu2019mo2cap2,tome2019xr,rhodin2016egocap,zhao2021egoglass,hakada2022unrealego}, several works show the potential of eyeglasses with two binocular fisheye cameras~\cite{zhao2021egoglass}. EgoCap~\cite{rhodin2016egocap} is one of the first works that adopt a helmet form factor with two mounted cameras on an extended stick. EgoGlass~\cite{zhao2021egoglass} extends the work to handle self-occlusion cases and 
UnrealEgo~\cite{hakada2022unrealego} improves the accuracy by using a large-scale synthetic dataset. 
The accuracy, however, is still far below third-person-view-based 3D pose estimation techniques. 

\begin{figure}[pt]
    \includegraphics[scale=0.2]{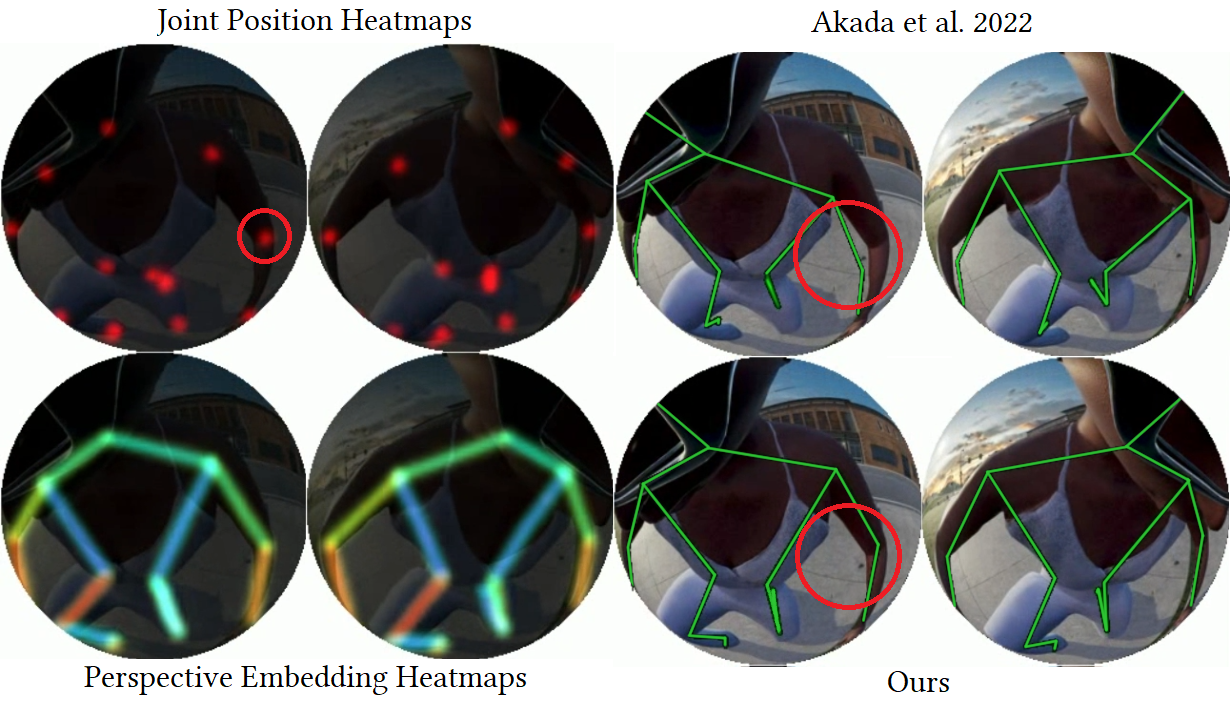}
    \caption{Heatmaps (Left) and 3D pose reprojected to the input images (Right) of \cite{hakada2022unrealego} and our methon. The heatmap estimation indicates that the visual information is sufficient.
    }
    \label{fig:reproj_error_sample}
\end{figure}

In this work, we introduce {\pp}, a novel system designed with an in-depth investigation of the binocular and egocentric setting. Compared to third-person view inputs, egocentric binocular inputs inherently possess limited information due to their restricted field of view and significant self-occlusions among the joints. Yet, existing approaches have not fully harnessed the potential of these already constrained inputs. Figure~\ref{fig:reproj_error_sample} shows that existing work fails on joints clearly shown in the inputs with sufficient visual information. This is especially common in motions with highly independent body parts movement, as shown in the supplementary video, where full-body information barely helps estimate specific limb poses.

We identify two under-utilized sources of 3D information within binocular and egocentric inputs: \textbf{stereo correspondence} and \textbf{perspective}. While these concepts are not new, we recognize that effectively incorporating them into the network poses unique challenges and requires careful investigation.

Utilizing stereo correspondence is challenging in two-fold. Firstly, existing methods in the domain of multiview pose estimation are highly data-dependent. Their performance lacks generalizability across poses or camera configurations~\cite{generalizable2022}. This is, in particular, challenging in egocentric settings with limited datasets. Secondly, 2D-to-3D lifting is inevitably ambiguous unless both 2D joint positions and camera calibrations are accurate~\cite{2dposematching2017}. Existing methods focus on optimizing these two components working sequentially. In contrast to third-person-view settings, where high accuracy can be achieved in 2D pose estimation, egocentric settings suffer from severe self-occlusions and view distortions propagating the error to stereo correspondence estimation.
Thus, tighter cooperation between stereo information and the 2D pose estimation holds greater significance.

The amplified perspective effect of fisheye camera views, especially with the egocentric setup, has been considered a challenge rather than an opportunity. Specifically, the body parts further from the camera, such as the lower limb, are shown much smaller than the real size, while the shoulder and upper limb views are enlarged.
Existing methods aim to tackle the challenge with reprojection methods to alleviate the distortion effect ~\cite{xu2019mo2cap2,egofish3d2023,autocalib2021} or extensively train the network with large synthetic datasets~\cite{tome2019xr,hakada2022unrealego}. However, these approaches overlook the potential value of the perspective factor, which can be useful information conveying the intensity of the 3D effect.

We propose two novel modules integrated into an end-to-end neural network to tackle the challenges.
First, we devise a Stereo Matcher network to serve two purposes: it i) focuses on learning the stereo correspondences independent from the full body pose distribution in the dataset and ii) generates an explicit estimation of the 3D orientation instead of directly lifting 2D joint estimations to a final 3D position. Specifically, the Stereo Matcher takes the optical features of each limb independently and predicts per-limb 3D orientation with a weight-shared network. Using the stereo correspondences and perspective effect, the architecture enables the module to focus on estimating the limb’s orientation without reliance on the full-body information. Furthermore, the estimated 3D orientation is concatenated with the 2D heatmaps' encoded pose features to assist lifting from 2D to 3D poses.

We also introduce the Perspective Embedding Heatmap representation.
This representation allows the 2D module to explicitly extract and utilize the 3D information of the perspective effect. We represent the perspective effect as the 3D orientation of the limbs.
For example, as shown in Figure~\ref{fig:reproj_error_sample}, the width of the top part of the upper limb is much larger than the width of the lower part. This visual cue suggests that the upper limb is nearly perpendicular to the camera view plane.
Then, we devise a trigonometry-based representation of the 3D orientation. It allows embedding the information into a 2D heatmap to utilize its uncertainty estimates.

Our end-to-end network, which integrates the two modules, significantly outperforms prior works. The quantitative measure of 3D pose estimation errors, MPJPE, was improved by 27.0\% and 23.1\% compared to the baselines EgoGlass and UnrealEgo in the UnrealEgo dataset, respectively. Furthermore, the qualitative results shown in Figure~\ref{fig:teaser} and Figure~\ref{fig:qualitative} also show noticeable improvements made by our work.

In summary, the main contributions are as follows:
\begin{itemize}
\item{We identify two under-utilized but crucial 3D information from binocular egocentric 3D pose estimation. }
\item{We propose a novel two-path network that separates stereo correspondence features from the pose features.}
\item{We introduce a new heatmap representation for 3D orientation to leverage the perspective effect of egocentric views.}
\item{We achieve significant quantitative and qualitative performance improvements compared to state-of-the-art methods.}
\end{itemize}

%% file: 03_related_work.tex
\section{Related Works}

\begin{figure*}[pt]
    \includegraphics[scale=0.47]{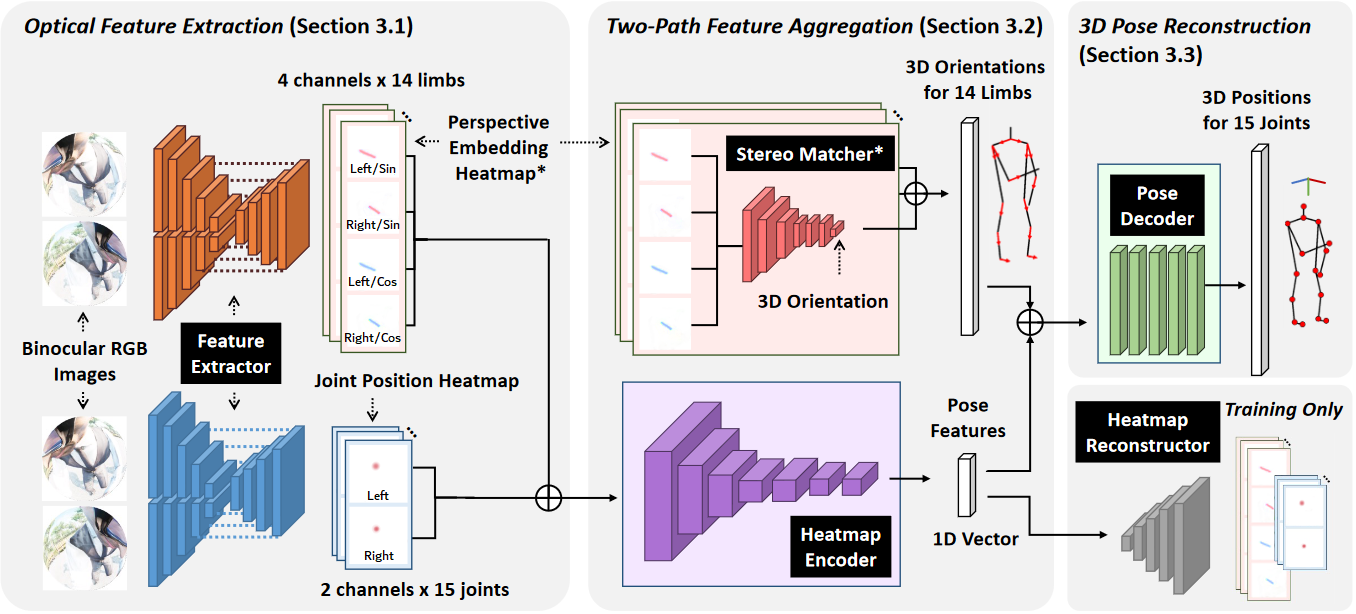}
    \caption{Our system consists of three stages: optical feature extraction, two-path feature aggregation, and 3D pose reconstruction with our two key components Perspective Embedding Heatmap and Stereo Matcher (marked with asterisk*). The optical feature extraction uses two feature extractors to generate Joint Position Heatmaps and Perspective Embedding Heatmaps.
    The Stereo Matcher network estimates the 3D orientation and learns the stereo correspondence from Perspective Embedding Heatmaps. The heatmap encoder takes all heatmaps as input and generates pose features. Finally, the 3D pose decoder takes the estimated 3D orientations and pose features to predict the 3D pose of the joints. The heatmap reconstructor is only used during training.}
    \label{fig:architecture}
\end{figure*}

\subsection{Binocular Egocentric 3D Pose Reconstruction}
There has been a growing body of literature on egocentric 3D pose reconstruction, particularly in the context of binocular vision-based setups. Compared to monocular vision-based methods that use a single fisheye camera with a downward view towards the full body~\cite{xu2019mo2cap2, tome2019xr}, binocular vision allows higher accuracy in 3D pose reconstruction via the wider view with cameras in proximity to the face~\cite{zhao2021egoglass}.

EgoCap~\cite{rhodin2016egocap} proposes the first binocular fisheye camera-based prototype mounted on a helmet or a headset. It uses the prototype to collect data with 8 subjects in a multiview studio setting for ground truth collection and creates a CNN-based 3D pose estimation model. EgoGlass~\cite{zhao2021egoglass} proposes compact binocular cameras embedded in eyeglasses, which are located in closer proximity to the user's body. However, the compact form factor leads to frequent self-occlusions of partial body parts and limited body coverage for various motions. The work partially solves the problem by explicitly forcing the model to learn the body part information with a segmentation mask. 
UnrealEgo~\cite{hakada2022unrealego} recently released a large-scale synthesized dataset for a similar setup of EgoGlass. It also shows that the overall 3D pose estimation accuracy improves when the weights between the two heatmap estimators, one for each of the binocular inputs, are shared. Despite the potential of such a setting, the overall pose estimation accuracy still requires significant improvement.

\subsection{Methods for Robust 3D Pose Estimation}
3D pose estimation solutions can be broadly divided into two types: i) direct estimation of 3D poses~\cite{maxmargin2015,pavlakos17volumetric, pavlakos20183dpose, Li20143DHP, Sun2017CompositionalHP, structuredpred2016} and ii) leveraging robust 2D pose estimation as an intermediate result and lifting to 3D~\cite{chen20173dpose,Jain2013LearningHP,MorenoNoguer20163DHP,conf/cvpr/TompsonGJLB15,li2019generating,hemlets2022,crossview2019,Tekin2016LearningTF}. The 2D-to-3D lifting method is known to be more robust in various situations~\cite{chen20173dpose}. In particular, maintaining the probabilistic nature of the intermediate 2D representation significantly contributed to performance improvement. Earlier work designed the 2D module to make multiple pose predictions to have ambiguity~\cite{Jahangiri20173dpose,moreno20173dpose}. Currently, a common approach is representing the estimated 2D joint positions as heatmaps~\cite{conf/cvpr/TompsonGJLB15, hemlets2022, Tekin2016LearningTF}. This approach has been explored in various ways. In particular, using skeletal heatmaps instead of joint heatmaps was proposed to handle occlusion cases in multi-person 3D pose estimation~\cite{8451055}. The key idea was to represent the skeleton as lines connecting two joints. We also leverage the line-based representation but propose a trigonometry-based embedding showing better performance. The heatmap-based approach was explored in multiview settings as well~\cite{crossview2019, zhang2020adafuse,iskakov2019learnable}. We also leverage the heatmap representations but propose methods to integrate 3D supervision from earlier stages of the 2D module.

Most of the recent works in egocentric 3D pose reconstruction also adopt the heatmap-based two-staged approach. $x$R-EgoPose \cite{tome2019xr}, which relies on a monocular fisheye view, uses a 2D module to generate Joint Position Heatmaps. Then, the 3D pose estimator module uses the heatmap as input to estimate the 3D pose. EgoGlass~\cite{zhao2021egoglass} uses the same architecture but extends the 2D module to take binocular inputs and estimates an additional segmentation mask for body part information. UnrealEgo~\cite{hakada2022unrealego} improves this work by using a weight-shared encoder of the U-Net for the two binocular inputs and a shared decoder for the heatmap estimation. Compared to the large body of work in third-person view 3D pose estimation, more studies are required to increase the robustness in egocentric settings.

%% file: 04_method.tex
\section{Method}
To better highlight the effectiveness of our two modules, incorporating stereo correspondence and perspective effect, we integrate them into widely used network architecture instead of redesigning the end-to-end network. In this paper, the overall structure follows the encoder-decoder network, where the 2D encoder module extracts optical features from the binocular RGB inputs as 2D Joint Position Heatmaps. The heatmaps are encoded into pose features and then reconstructed into 3D joint positions with the final 3D decoder. Here, we add the Perspective Embedding Heatmap generator in the optical feature extraction stage and the Stereo Matcher to the encoder stage.

Figure~\ref{fig:architecture} shows the overall network architecture. The input to the network is two 256$\times$256 RGB images. The first stage, optical feature extraction, generates two sets of heatmaps: i) Perspective Embedding Heatmaps based on our novel perspective-aware trigonometry representation and ii) Joint Position Heatmaps similar to previous work. Then, the two-path feature aggregation network separately learns i) the stereo correspondences from the binocular Perspective Embedding Heatmaps and ii) pose feature vector extracted from both Joint Position Heatmaps and Perspective Embedding Heatmaps. Our network then concatenates the two disentangled information and generates the final 3D local pose. The local pose is defined as a 3D position of the joints relative to the pelvis in the camera coordinate system. A total of 15 joints are estimated, including the neck, shoulders, elbows, hands, thighs, calves, ankles, and feet. We additionally train a heatmap reconstruction network which is not used for inferences.

\subsection{Optical Feature Extraction}

\begin{figure}
    \centering
    \includegraphics[width=0.9\linewidth]{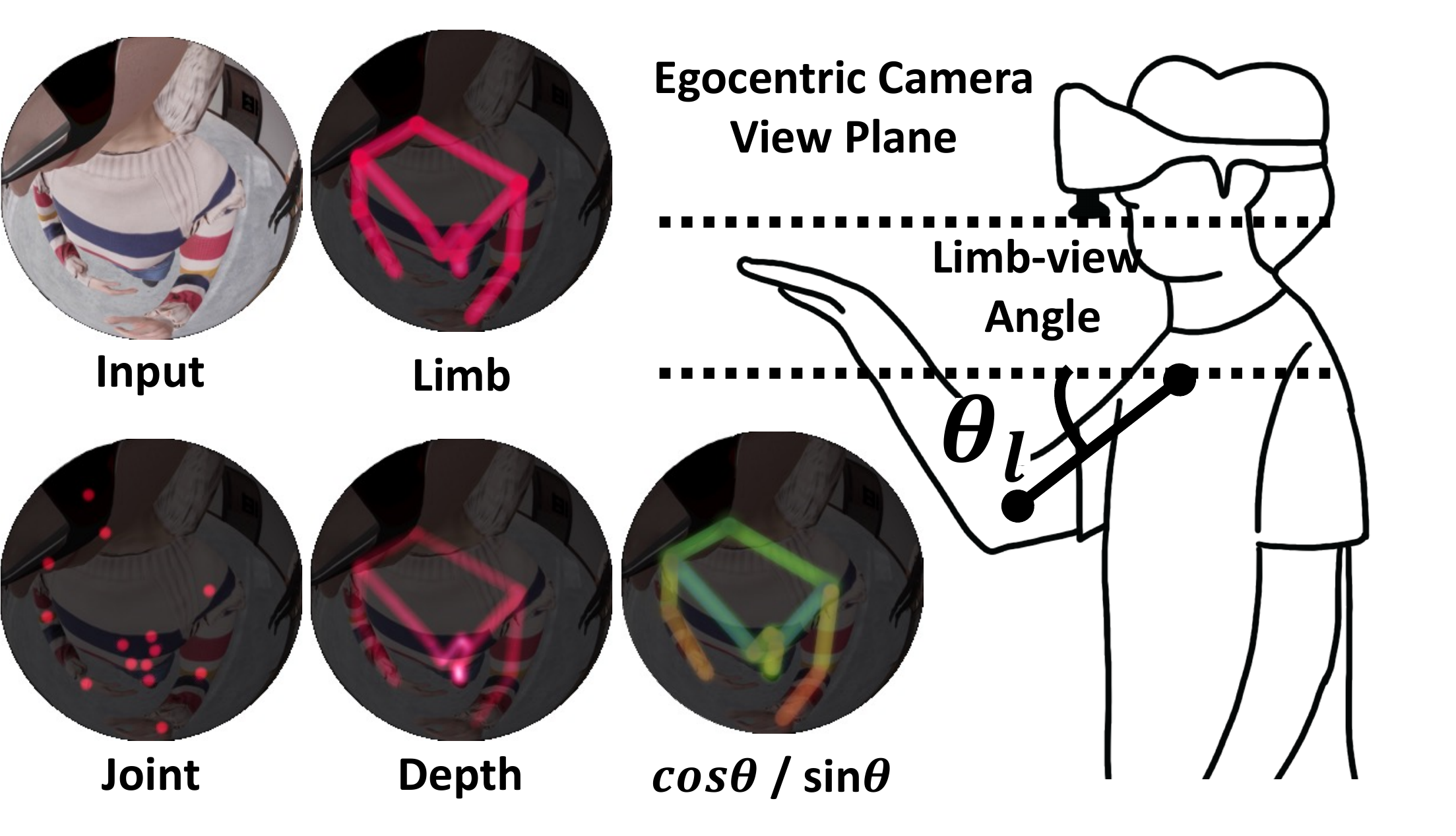}
    \caption{Visualization of a joint position, limb position, limb depth, and our Perspective Embedding Heatmap, where the color represents the cosine and sine of the limb-view angle shown in the right image.}
    \label{fig:heatmap}
\end{figure}

This module consists of two networks. Both networks take the binocular RGB images as input but generate different types of optical features: Perspective Embedding Heatmap and Joint Position Heatmap. They use the same network architecture, a U-Net-based model with a ResNet18 backbone. 
We first explain the Joint Position Heatmap as a preliminary for our Perspective Embedding Heatmap.

\subsubsection{Joint Position Heatmap}
This heatmap is a single-channel image where each pixel represents the confidence level associated with the 2D positions of the joints. It has been widely adopted in various pose estimation methods~\cite{10.1007/978-3-319-46478-7_44} due to its effectiveness in capturing uncertainty.
To align with previous work, we adopt a per-joint heatmap generation approach instead of using a single heatmap for all joints. A total of 30 Joint Position Heatmaps are generated, corresponding to the left and right input images for the 15 individual joints.

\subsubsection{Perspective Embedding Heatmap} \label{rot_heatmap}
The goal of the Perspective Embedding Heatmap is to incorporate the probabilistic information of the perspective effect instead of the joint positions. The perspective effect can be expressed in different ways, such as the depth of each joint or the angle between two adjacent joints (Refer to Figure~\ref{fig:heatmap} for the visualization of different types and the specific explanations in Section~\ref{subsubsec:effectiveness-heatmap}). We, in particular, choose the 3D orientation of the limbs defined with the angle between the viewing plane and the limb (referred to as limb-view angle hereinafter). The limb-view angle $\theta_l$ for each limb is computed as follows:

$$\theta_l = \text{atan}\left(\frac{z_l}{\sqrt{x_l^2 + y_l^2}}\right)$$

\noindent{}, where $x_l, y_l, z_l$ denotes the relative 3D coordinate of the child joint from the parent joint in the camera coordinate space, and $xy$-plane is equivalent to the viewing plane of the camera. 
Limbs are represented as a line rather than a pair of two adjacent joints to enhance the robustness against occlusions because the visible parts of the limbs can be utilized.

Embedding the 3D orientation to a heatmap is not straightforward. We have to embed three pieces of information, 2D position, confidence, and 3D orientation, into a 2D heatmap.
One approach would be simply multiplying the magnitude of the orientation with the probabilistic output of the heatmap. However, this can confuse the network as the pixel values contain information on confidence and orientation. Adding the perspective value as a separate channel is also not an effective representation because the confidence only refers to the 2D joint position estimation rather than the orientation value.

We use the trigonometric function to embed the limb-view angle into a 2D heatmap. This choice is motivated by the properties of trigonometric functions. The sine and cosine value represents an angle with the constant norm as a vector. 
Our heatmap has two channels where the pixel values are scaled sine and cosine of the limb-view angle. The scale of the resulting vector represents the confidence of the orientation estimation, while the ratio of the sine and cosine values can represent the estimated limb-view angle. In this way, using the trigonometric functions ensures that the perspective information is entangled with the probability.
Perspective Embedding Heatmaps are defined for 14 limbs, excluding the head-neck connection from the 15 limbs connecting the tree hierarchy of 16 joints. Each set of Perspective Embedding Heatmaps contains four channels of heatmaps for sine and cosine for the left and right camera views.

\subsection{Two-Path Feature Aggregation}\label{subsec:stereo-correspondence}
The key to our high 3D pose estimation accuracy is devising the network to learn the stereo feature matching separately from the distribution of full body pose in the train dataset. Thus, we replace the one-path encoder stage of the previous work, which generates pose features from the 2D joint heatmaps, with a two-path feature aggregation stage. We first explain the heatmap encoder network, which extends prior work estimating the pose features. Then, we provide the details of the new Stereo Matcher network.

\subsubsection{Heatmap Encoder}\label{subsubsec: heatmap-encoder}
The input of the heatmap encoder is all the heatmaps generated from the previous optical feature extraction stage. The heatmaps include the joint positions of all 15 joints and Perspective Embedding Heatmaps for 14 limbs concatenated channel-wise. The heatmap encoder is a CNN-based network, and the output is a 1D vector of 20 pose feature values.

\subsubsection{Stereo Matcher Network}\label{subsubsec: stereo-matcher}
Our Stereo Matcher network is a simple yet effective technique that takes the Perspective Embedding Heatmaps of each limb as input and estimates the 3D orientation of the limb individually. The 3D orientation of each limb is defined as,

$$\mathbf{o}_l = \frac{{[\mathbf{x}_l, \mathbf{y}_l, \mathbf{z}_l}]}{{\sqrt{{\mathbf{x}_l^2 + \mathbf{y}_l^2 + \mathbf{z}_l^2}}}}
$$

\noindent{}, a normalized vector of relative positions of a child from the parent joint in a camera's coordinate system. $x_l, y_l, z_l$ denotes the relative 3D coordinate.

\aptLtoX{\begin{figure}[t]
     \includegraphics[width=.49\textwidth]{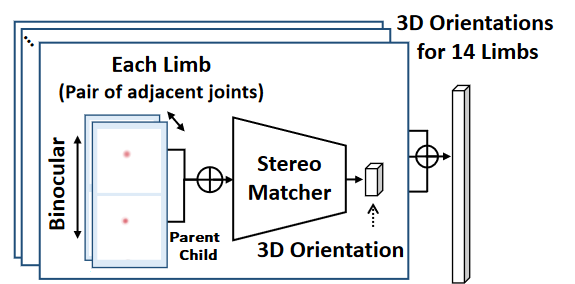}
     \caption{Our Stereo Matcher module without Perspective Embedding Heatmap.}\label{fig:ours-sm}
 \end{figure}
\begin{figure}
  \centering
     \includegraphics[width=.49\textwidth]{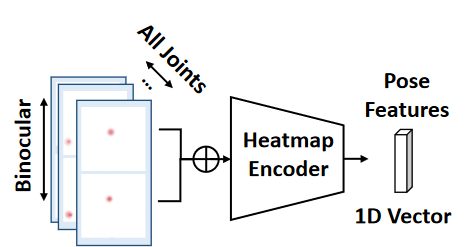}
     \caption{Pose feature extraction module from prior work.}\label{fig:baseline-heatmap-encoder}
\end{figure}}{\begin{figure}[t]
\begin{minipage}[t]{0.49\linewidth}
     \centering
     \includegraphics[width=1.0\linewidth]{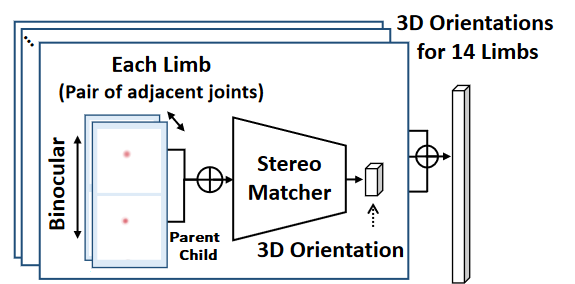}
\captionsetup{width=0.9\linewidth}    
     \caption{Our Stereo Matcher module without Perspective Embedding Heatmap.}\label{fig:ours-sm}
 \end{minipage}\hfill
\begin{minipage}[t]{0.49\linewidth}
  \centering
     \includegraphics[width=1.0\linewidth]{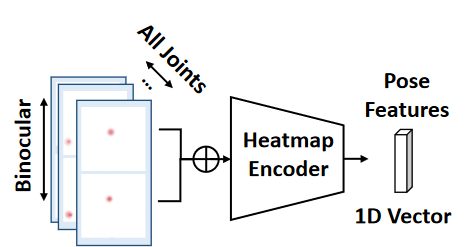}
     \captionsetup{width=0.9\linewidth}          
     \caption{Pose feature extraction module from prior work.}\label{fig:baseline-heatmap-encoder}
\end{minipage}\hfill
\end{figure}}

The Stereo Matcher takes the heatmaps of one limb at a time and estimates the specific limb's 3D orientation. Figure~\ref{fig:ours-sm} visualizes the architecture, with the conventional heatmaps used for comparison. Each limb is represented as a pair of position heatmaps of adjacent joints (i.e., a total of 4 channel heatmaps, including binocular Joint Position Heatmaps of two joints). The per-limb process is clearly distinguished from the heatmap encoder shown in Figure~\ref{fig:baseline-heatmap-encoder}, which takes heatmaps for all body parts to encode them as one pose feature vector. The estimation without full body information is possible due to stereo information and 3D cues from Perspective Embedding Heatmaps, and it prevents reliance on the trained full-body pose distribution. The process does not require specific knowledge of the limb. Thus, the weights of the Stereo Matcher network are shared across different limbs for sample efficiency. Our final network architecture integrates the Stereo Matcher with the Perspective Embedding Heatmaps shown in Figure~\ref{fig:architecture}. The four channels input is the Perspective Embedding Heatmaps of one limb, including two channels each for sine and cosine. Our empirical analysis shows that the integration has a synergistic effect (Table~\ref{table:performance_breakdown}).

\subsection{3D Pose Reconstruction}\label{subsubsec:end-to-end}
In the final 3D pose reconstruction stage, the 3D pose decoder takes the concatenated vector of the pose features and the 3D orientations of 14 limbs as input. Then, it estimates the final 3D local pose of the 15 joints. The 3D pose decoder consists of only fully connected layers. Lastly, our system adopts a heatmap reconstructor of \cite{tome2019xr}, only used during training, to constrain the heatmap encoder to preserve the uncertainty represented in the heatmap.

%% file: 05_training.tex
\section{Training Details}
\label{sec:training-details}
 Our \pp{} is trained and evaluated on a cloud server with Nvidia RTX A6000 48GB GPU, and two AMD EPYC 7313 CPUs. The entire process of training takes approximately 20 hours on the UnrealEgo dataset. The network is trained in 2 stages, first for only an optical feature extractor, and second for only a 3D pose reconstructor. For the training of the 3D pose reconstructor, pre-trained and frozen optical feature extractor weights are used to estimate heatmaps.
 
The ResNet-18 in the U-Nets of the optical feature extractors are initialized with pre-trained weights. Other networks are initialized with standard normal distribution. The 3D orientation estimated from Stereo Matcher (Section \ref{subsubsec: stereo-matcher}) is detached from the gradient graph before being used as input to a 3D pose estimator to ensure that the module is learned independently.

 The network is trained for 10 epochs, with linear learning rate decay from epochs 5 to 10 on the UnrealEgo dataset. For the EgoCap dataset, due to its volume, the network is trained for 100 epochs. Batch size 16 is used. Adam~\cite{KingBa15} optimizer with learning rate $10^{-3}$ is used for the optical feature extractor. The learning rate-free~D-Adaptation optimizer~\cite{defazio2023learningratefree}, D-Adapt SGD is used for the 3D pose reconstruction to facilitate the tuning process with a growth rate of $1.02$. Empirically, the SGD variant is more effective with the intermediate output detached during the training. More details of the network architecture are described in the supplementary material.

\subsection{Loss}
We introduce the details of the loss term used for training our system. We train modules in 2 stages, and multiple losses are defined for each stage.

\subsubsection{Optical Feature Extractor}
Denoting the estimated Joint Position Heatmaps as $\hat{JH}$ and the ground truth heatmaps as $JH$, the loss for the Joint Position Heatmap estimation is computed as follows:
\begin{equation*}
L_{JH} = mse(\hat{JH}, JH) 
\end{equation*}

To encourage learning the heatmap evenly for limbs with different pixel lengths, the Perspective Embedding Heatmap estimation loss for each limb is divided by its pixel length, accounting for the estimation error dependency on the length of the limb in the pixel space. We observed that this helps to get stable test accuracy. For the Perspective Embedding Heatmaps, denoting the estimated heatmaps as $\hat{PH}$ and the ground truth heatmaps as $PH$, the loss is computed as follows:

\begin{equation*}
L_{PH} = \frac{1}{\lvert S \rvert}\sum_{l \in S}\frac{1}{\lvert \ell_l \rvert}\text{mse}(\hat{PH_l}, PH_l)
\end{equation*}

$PH_l$ denotes a set of heatmaps for limbs, ${\ell_l}$ is the pixel length of the line in the ground truth heatmap, connecting two projected coordinates of the ends of the limb, and $S$ is the set of limbs used for the Perspective Embedding Heatmaps.

Loss weights for the optical feature extractor are set to $\lambda_{JH} = 1$ and $\lambda_{PH} = 1$. The total loss ($L_{2D}$) for the  optical feature extractor is the following:
\begin{equation*}
L_{2D} = \lambda_{JH} * L_{JH} + \lambda_{PH} * L_{PH}
\end{equation*}

\subsubsection{3D Pose Reconstructor}
The loss for the Stereo Matcher is defined as follows, 

\begin{equation*}
L_{Trans} = \text{mse}(\hat{\mathbf{o}}, \mathbf{o})
\end{equation*}

\noindent{}, where $\mathbf{o}$ is the ground truth of all limbs' 3D orientation and $\hat{\mathbf{o}}$ is an estimated orientation of all limbs.

The pose estimator is trained with 3D pose reconstruction loss, heatmaps reconstruction loss, and cosine-similarity loss~\cite{hakada2022unrealego} as follows:

\begin{equation*}
L_{Pose} = \Sigma_{j \in J} ||\hat{p_j}-p_j||
\end{equation*}
\begin{equation*}
L_{Recon} = (\text{mse}(\hat{JH}, \tilde{JH}) + \text{mse}(\hat{PH}, \tilde{PH}))
\end{equation*}
\begin{equation*}
L_{Cos} = \sum_{l \in S} \frac{f(l, \hat{p}_j) \cdot f(l, p_j)}{||f(l, \hat{p}_j)|| ||f(l, p_j)||}
\end{equation*}

\noindent{}, where $J$ is the set of joints to estimate 3D position. $f(l, p_j)$ is a limb $l$'s translation, the child joint's 3D coordinate relative to the parent joint, computed from given joint positions $p_j$. $S$ is the set of limbs used by our systems.
We set loss weights of $\lambda_{Trans} = 1$, $\lambda_{Pose} = 10^{-1}$, $\lambda_{Recon} = 10^{-3}$, and $\lambda_{Cos} = -10^{-2}$ for the 3D pose reconstructor. The total loss ($L_{3D}$) for the 3D pose reconstructor is the following:
\begin{equation*}
L_{3D} = \lambda_{Trans} * L_{Trans} + \lambda_{Pose} * L_{Pose} + \lambda_{Recon} * L_{Recon} + \lambda_{Cos} * L_{Cos}
\end{equation*}

\renewcommand{\tabcolsep}{1pt}
\begin{table*}[t]
\small
\centering
\begin{center}
 \caption{Comparison with the state-of-the-art binocular egocentric pose reconstruction methods on the \textbf{UnrealEgo} dataset.}
\begin{tabular}{ |p{1.75cm}||p{1.75cm}|p{1.75cm}|p{1.75cm}|p{1.75cm}|p{1.75cm}|p{1.75cm}|p{1.75cm}|p{1.75cm}|  }
 \hline
 \multicolumn{9}{|c|}{MPJPE(PA-MPJPE)} \\
 \hline
 Method & \textbf{Overall} & Jumping & Falling Down & Exercising & Pulling & Singing & Rolling & Crawling \\
 \hline
 EgoGlass & 83.33(61.56) & 81.33(63.42) & 131.64(94.70) & 100.17(74.11) & 81.96(62.07) & 70.59(52.57) & 103.29(92.66) & 182.36(113.65) \\
 UnrealEgo & 79.08(59.26) & 76.63(61.13) & 126.82(95.99) & 90.27(68.36) & 78.20(61.39) & 67.34(49.85) & 86.26(71.61) & 181.50(116.42) \\
 Ours & \textbf{60.82(48.47)} & \textbf{60.84(49.41)} & \textbf{95.78(79.17)} & \textbf{76.15(63.31)} & \textbf{58.08(45.69)} & \textbf{51.75(41.05)} & \textbf{81.94(71.16)} & \textbf{148.72(105.14)} \\ 
\end{tabular}
\begin{tabular}{ |p{1.75cm}||p{1.75cm}|p{1.75cm}|p{1.75cm}|p{1.75cm}|p{1.75cm}|p{1.75cm}|p{1.75cm}|p{1.75cm}|  }
\hline
 \multicolumn{9}{|c|}{MPJPE(PA-MPJPE)} \\
 \hline
Method & Laying & Sitting on the Ground & Crouching & Crouching and turning & Crouching to Standing & Crouching-Forward & Crouching-Backward & Crouching-Sideways \\
 \hline
 EgoGlass & 109.64(81.75) & 207.31(144.72) & 132.74(110.83) & 128.50(100.83) & 82.30(60.48) & 79.23(65.41) & 94.77(77.98) & 93.89(77.91) \\
 UnrealEgo & 97.84(76.62) & 194.34(150.33) & 116.77(97.88) & 128.23(104.32) & 76.80(55.18) & 73.60(61.89) & 78.19(62.01) & 84.91(71.41) \\
 Ours & \textbf{83.36(69.85)} & \textbf{153.81(133.37)} & \textbf{93.59(79.64)} & \textbf{109.05(89.97)} & \textbf{65.34(48.70)} & \textbf{53.99(45.94)} & \textbf{58.13(46.90)} & \textbf{67.67(56.69)} \\
\end{tabular}
\begin{tabular}{ |p{1.75cm}||p{2.01cm}|p{2.01cm}|p{2.01cm}|p{2.01cm}|p{2.01cm}|p{2.01cm}|p{2.0cm}|  }
 \hline
  \multicolumn{8}{|c|}{MPJPE(PA-MPJPE)} \\
 \hline
Method & Standing-Whole Body & Standing-Upper Body & Standing-Turning & Standing to Crouching & Standing-Forward & Standing-Backward & Standing-Sideways \\
 \hline
 EgoGlass & 70.17(51.55) & 70.18(46.32) & 77.72(59.29) & 77.82(70.02) & 77.34(62.27) & 75.13(58.47) & 85.19(67.76) \\
 UnrealEgo & 68.47(50.10) & 66.54(45.44) & 74.33(57.92) & 74.09(61.25) & 70.79(56.61) & 68.13(55.17) & 78.92(62.27) \\
 Ours & \textbf{50.37(39.90)} & \textbf{50.51(37.19)} & \textbf{58.52(48.08)} & \textbf{59.64(52.78)} & \textbf{55.71(48.72)} & \textbf{51.96(42.58)} & \textbf{60.93(51.22)} \\
\end{tabular}
\begin{tabular}{ |p{1.75cm}||p{1.75cm}|p{1.75cm}|p{1.75cm}|p{1.75cm}|p{1.75cm}|p{1.75cm}|p{1.75cm}|p{1.75cm}|  }
 \hline
  \multicolumn{9}{|c|}{MPJPE(PA-MPJPE)} \\
 \hline
Method & Dancing & Boxing & Wrestling & Soccer & Baseball & Basketball & Football & Golf \\
 \hline
 EgoGlass & 83.72(65.50) & 71.38(54.25) & 86.09(65.28) & 82.29(58.24) & 77.69(61.04) & 59.68(47.73) & 102.15(80.56) & 69.37(48.22) \\
 UnrealEgo & 79.99(63.20) & 68.69(51.39) & 83.73(64.47) & 78.32(56.24) & 72.81(56.73) & 60.48(45.41) & 99.56(81.81) & 73.86(49.01) \\
 Ours & \textbf{61.30(50.82)} & \textbf{49.91(40.25)} & \textbf{65.48(51.76)} & \textbf{54.46(43.70)} & \textbf{69.25(55.58)} & \textbf{48.39(35.83)} & \textbf{83.28(72.96)} & \textbf{54.52(38.10)} \\
 \hline
\end{tabular}
 \label{comparison}
 \end{center}
\end{table*}

%% file: 06_evaluation.tex
\section{Evaluation}

\begin{table}
\caption{Overall error comparison of \pp, EgoGlass and UnrealEgo on real-world \textbf{EgoCap} dataset. }
\begin{tabular}{ |c||m{1.6cm}|m{1.6cm}|m{1.6cm}| }
 \hline
\begin{imageonly}\diagbox{Error}{Methods}\end{imageonly} & EgoGlass  & UnrealEgo & \textbf{Ours} \\
 \hline
 MPJPE & 61.78 & 59.16 & \textbf{54.41} \\
  PA-MPJPE & 46.06 & 48.22 & \textbf{40.24} \\
 \hline
\end{tabular}
\label{table:egocap}
\end{table}

\begin{table}
\caption{Ablation results of Perspective Embedding Heatmap (\textbf{PH}), Stereo Matcher (\textbf{SM}) on \textbf{UnrealEgo} Dataset. The input for \textbf{B+SM} evaluation is two adjacent joint heatmaps.}
\begin{tabular}{ |c||m{1.6cm}|m{1.2cm}|m{1.2cm}|m{1.6cm}| }
 \hline
\begin{imageonly} \diagbox{Error}{Methods}\end{imageonly} & Baseline(\textbf{B}) & \textbf{B+PH} &\textbf{B+SM}&\textbf{B+PH+SM}  \\
 \hline
 MPJPE & 79.08 &75.82&66.72&\textbf{60.82} \\
  PA-MPJPE  & 59.26 & 58.52 & 52.29 & \textbf{48.47} \\
 \hline
\end{tabular}
 \label{table:performance_breakdown}
\end{table}

\begin{table}
 \caption{Ablation of two-path feature aggregation on \textbf{UnrealEgo} dataset. The Heatmap Encoder (\textbf{HE}) module is used exclusively, and then with our Stereo Matcher (\textbf{SM}), Rotation Decoder\cite{tome2020self} (\textbf{RD}), and Orientation Decoder (\textbf{OD}).}
\begin{tabular}{ |c||m{1.08cm}|m{1.08cm}|m{1.08cm}|m{1.08cm}|m{1.08cm}| }
 \hline
\begin{imageonly} \diagbox{Error}{Paths}\end{imageonly} & HE & HE+RD & HE+OD & HE+SM & SM \\
 \hline
 MPJPE & 75.82 & 73.48&70.77&\textbf{60.82} & 73.40 \\
 PA-MPJPE & 58.52 & 57.32 & 55.83 &\textbf{48.47} & 51.62 \\
 \hline
\end{tabular}
 \label{table:rot_decoder}
\end{table}
\begin{table}
\centering
 \caption{Comparison of results on \textbf{UnrealEgo} dataset with different representations used for Perspective Embedding Heatmaps on the UnrealEgo dataset. Tri. is the result with the trigonometric function.}
\begin{tabular}{ |c||m{1.0cm}|m{1.0cm}||m{1.0cm}|m{1.0cm}|m{1.0cm}|m{1.0cm}| }
 \hline
\begin{imageonly}\diagbox{Error}{Heatmaps}\end{imageonly} & Joints & Line(\textbf{L}) & \textbf{L}\&Depth & \textbf{L}\&Angle & \textbf{L}\&Tri. \\
 \hline
 MPJPE & 66.72 & 65.66 &82.36& 65.68& \textbf{60.82}\\
 PA-MPJPE & 52.29 & 51.82& 63.41&52.27&\textbf{48.47}\\
 \hline
\end{tabular}
 \label{table:heatmap_types}
\end{table}

\subsection{Experimental Setup}
\subsubsection{Datasets} 
We evaluate using the publicly available synthetic UnrealEgo dataset, generated through the Unreal Engine~\footnote{\url{https://www.unrealengine.com}}. Moreover, we further demonstrate the practicality of our approach on the real-world EgoCap~\cite{rhodin2016egocap} dataset.
For our ablation studies, we primarily use the UnrealEgo dataset. It provides much more comprehensive coverage of diverse human movements across various settings, which surpasses that of the EgoCap dataset. Furthermore, it encompasses more challenging scenarios, particularly those involving severe self-occlusions. Lastly, the dataset covers various ethnicities and body types with 17 different characters.

\subsubsection{Baselines}
We evaluate our system against two state-of-the-art binocular egocentric pose estimation models, e.g., UnrealEgo~\cite{hakada2022unrealego} and EgoGlass~\cite{zhao2021egoglass}, using the evaluation dataset. The results for baselines are based on the provided open-source code~\cite{hakada2022unrealego}. The evaluation of UnrealEgo on its dataset is based on the open-source weight. To perform a comprehensive evaluation of EgoGlass, we re-implement its full architecture as the body part branch, which is the primary factor for its achieved performance, is not included in the source code. As the body part branch requires segmentation data not provided in the original UnrealEgo dataset, we utilize the pseudo-segmentation technique proposed by EgoGlass.

\subsubsection{Metrics}
We utilize the widely recognized Mean Per Joint Position Error (\textbf{MPJPE}) as a metric, which calculates the mean Euclidean distance between the predicted joint positions and their respective ground truth counterparts. Additionally, we also employ Procrustes Analysis MPJPE (\textbf{PA-MPJPE}) as a complementary metric to further quantify the precision and robustness of our approach.

\subsection{Overall Performance} 
\subsubsection{Evaluations on UnrealEgo Dataset} Table~\ref{comparison} shows the 3D pose estimation errors of our proposed system, \pp, and two baseline systems, e.g., EgoGlass and UnrealEgo, on the UnrealEgo dataset. Tests are conducted for all motion categories within the dataset, and our proposed system achieves the best results across the board. Specifically, our system demonstrates an MPJPE of only 60.82, which represents a significant improvement of 23.1\% and 27.0\% over the UnrealEgo and EgoGlass systems, respectively, in terms of overall error.
Our system outperforms previous methods in every motion category by a large margin, including challenging motions such as ``Crawling" and ``Sitting on the Ground".

\subsubsection{Evaluations on EgoCap Dataset} As the motions are not categorized in this dataset, we present the overall results. It should be noted that the EgoCap dataset solely provides 3D pose annotations for its test dataset publicly. We sorted the 3D test dataset by subject and frame, trained on 80\% of it, and reserved the remaining 20\% for testing purposes. Table \ref{table:egocap} shows that our method performs well in different settings and in real-world images.
Our approach manifests augmentations, as evidenced by 11.2\% improvements in MPJPE and 12.6\% enhancements in PA-MPJPE, in contrast to EgoGlass. Moreover, it outperforms UnrealEgo with 8.0\% advances in MPJPE and 16.5\% progressions in PA-MPJPE. 
It is worth noting that the EgoCap dataset exhibits fewer instances of severe self-occlusions and a narrower range of pose variations in comparison to the UnrealEgo dataset. The baseline methods achieve modest accuracy resulting in smaller performance improvement compared to the evaluation on the UnrealEgo dataset.

\subsubsection{Inference Time} 
The total time was 18ms making our network suitable for real-time applications. Our latency breakdown shows that the per-limb Stereo Matcher network incurs minimal computation overhead (e.g., Optical Feature Extraction: 13.88ms, Heatmap Encoder: 1.07ms, Stereo Matcher: 1.51ms, Pose Reconstruction: 0.32ms).
\cite{hakada2022unrealego} has shown that naively using a larger computational capacity, replacing the ResNet-18 backbone with up to the ResNet-101 backbone, does not improve accuracy. Our method effectively takes advantage of additional computation.

\subsubsection{Qualitative Comparison}
We demonstrate our system's accuracy qualitatively. Fig.~\ref{fig:qualitative} shows the comparison to the previous methods in both the UnrealEgo dataset and the EgoCap dataset. The results on the UnrealEgo dataset show that our method is significantly more accurate on challenging poses. The EgoCap dataset evaluation shows that our method performs much better compared to previous methods in the real-world setting. Refer to the video attached as supplementary material for more results.

\subsection{Ablation Studies}

\subsubsection{Breakdown of Accuracy Gains.} Table~\ref{table:performance_breakdown} shows the effectiveness of our two newly proposed techniques, Perspective Embedding Heatmap (PH) and Stereo Matcher (SM). The reduction of MPJPE is 4.1\% when only the Perspective Embedding Heatmap is used without the Stereo Matcher. Alternatively, with only the Stereo Matcher, using the joint position heatmaps as shown in Figure~\ref{fig:ours-sm}, a reduction of 15.6\% in MPJPE is achieved. Lastly, the combination of techniques results in a synergistic effect, leading to a substantial reduction of 23.1\% in MPJPE.

\subsubsection{Effectiveness of the Two-Path Feature Aggregation}
We now compare the effectiveness of our Stereo Matcher with design alternatives in Table~\ref{table:rot_decoder}. The Perspective Embedding Heatmap is used as an input to the Heatmap Encoder for all experiments. The two key features are i) per-limb 3D orientation estimation and ii) utilization of 3D orientation as an intermediate output to assist the Heatmap Encoder for the final 3d pose reconstruction.
Regarding the first feature, we evaluate it against a single network that predicts all the 3D orientations simultaneously from the concatenated heatmaps (referred to as Orientation Decoder in Table~\ref{table:rot_decoder}). This alternative only reduced the MPJPE by 6.6\%, while our network achieved an impressive 19.8\% improvement. We also show the effectiveness of the second feature by incorporating the rotation decoder proposed by \cite{tome2020self}. The decoder estimates 3D joint rotation from encoded pose features and guides the heatmap encoder to learn rotational features. This method is not as effective as using the 3D estimation as an input to the pose decoder.
The experimental result indicates that the inclusion of a rotation decoder yields only a minimal increase in accuracy for 3D pose recognition by only 3\%. Lastly, we evaluate using only the Stereo Matcher path in the last column. Stereo  Matcher alone outperforms the one with only the Heatmap Encoder, which shows the importance of the per-limb independent feature extraction.

\subsubsection{Effectiveness of Perspective Embedding Heatmap}
\label{subsubsec:effectiveness-heatmap}
We experimented with alternative representations to embed perspective information in the heatmaps. The result is shown in Table \ref{table:heatmap_types}.
We first compare joint and line-based representation. The line-based representation is a raw limb heatmap without perspective information. The heatmap value represents the position of the limb rather than the orientation, as in ours. The results show that limb representation has a slightly better performance compared to the two joint-based one. We also test the alternatives of our trigonometric-based representation of the perspective effect. Specifically, we compare two other alternatives: depth and raw angle. The pixel value of the depth heatmap is the depth of each point on the limb. We normalized and added an offset to the depth value to keep it larger than zero. The raw angle representation refers to the limb-view angle without applying the trigonometric functions. The results show that our trigonometric-based representation achieves the highest accuracy.

%% file: 07_conclusion.tex
\section{Conclusion}
In this work, we push the performance boundaries of binocular egocentric 3D pose estimation. We identify an under-utilized source of information in the highly constrained setting, stereo correspondence, and perspective effect. Our system incorporates a novel Stereo Matcher network, which learns to estimate 3D pose independently using stereo correspondence of heatmaps for each limb. It alleviates the issue of overfitting to the distribution of full body pose in the dataset. Furthermore, we introduce a 3D Perspective Embedding Heatmap representation. It effectively captures the perspective effect while preserving the inherent uncertainty associated with heatmaps. Our system demonstrates substantial improvements compared to state-of-the-art methods. It is noteworthy that our proposed techniques are not exclusively tailored to the egocentric and binocular setting. They hold potential for direct application in various contexts of human 3D pose reconstruction, providing benefits beyond the scope of this work.

\section*{Acknowledgement}
This work was supported by the National Research Foundation of Korea(NRF) grant funded by the Korea government(MIST) (No. 2022R1A2C3008495, No. RS-2023-00218601).

%% file: appendix.tex
\appendix
\begin{figure}[pt]
    \includegraphics[scale=0.1]{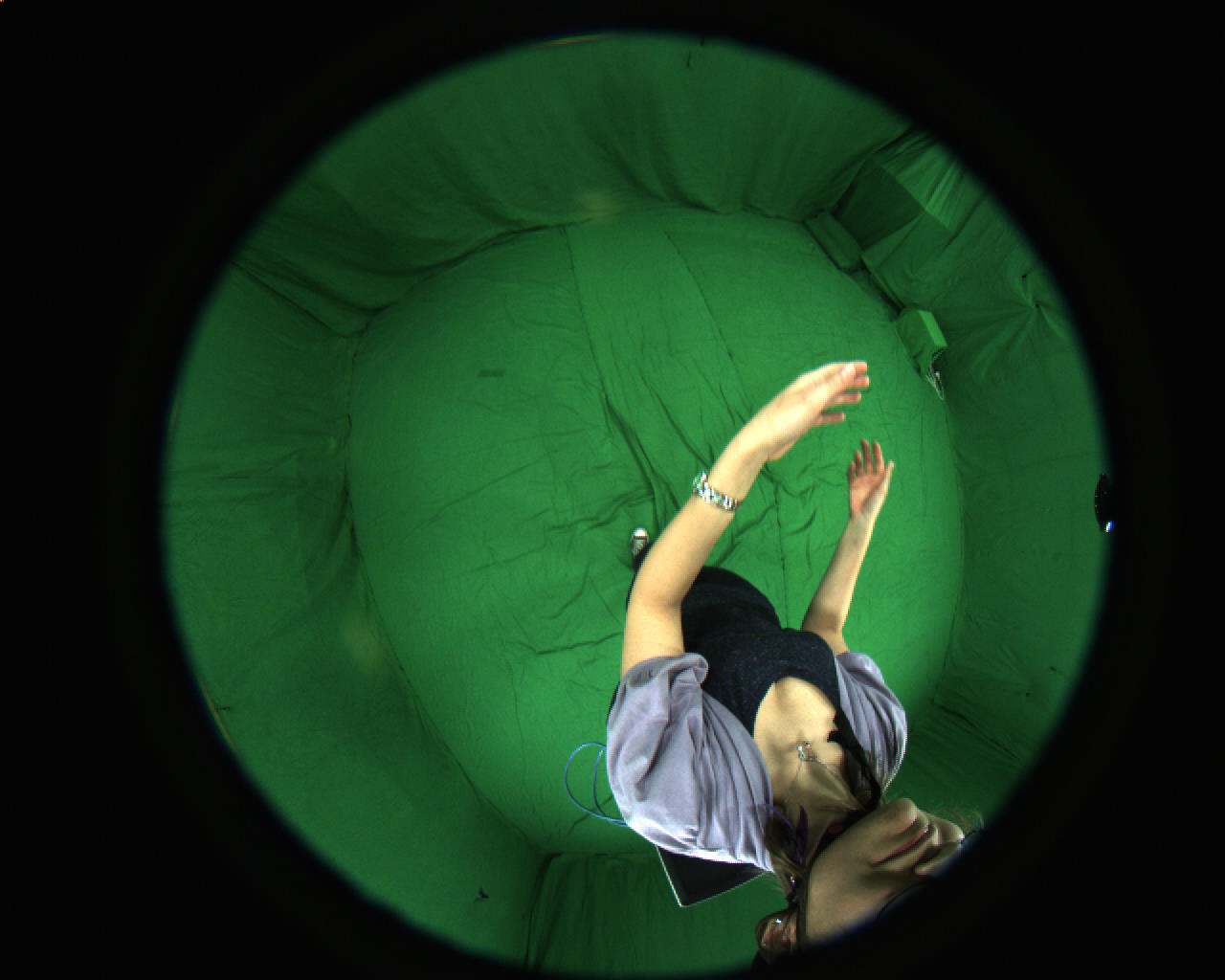}
    \caption{Sample of 1280$\times$1024 EgoCap dataset image.}
    \label{fig:egocapsample}
\end{figure}

\section{Data Preprocessing}
\subsection{UnrealEgo}
The full dataset of UnrealEgo~\cite{hakada2022unrealego} is utilized. We use the publicly available preprocessing, data loading, and training code. The dataset provides ground truth 2D and 3D poses, which we utilize to generate the ground truth for Perspective Embedding Heatmaps.

\subsection{EgoCap}
We only use the validation dataset for our evaluation. The 3D annotations for the training dataset are not publicly available. 

The validation set contains 2D and 3D versions. The 2D set contains all of the images in the 3D set. Thus, we use images from the 3D set, and the ground truth 2D joint position is obtained from the 2D set annotation, while the ground truth 3D pose is gathered from the 3D set.

The EgoCap dataset's original image dimension is 1280$\times$1024 as shown in Fig.\ref{fig:egocapsample}. A large portion of the image is empty. The full view of the camera approximately covers a circle of 512-pixel radius. In the prepossessing, images are horizontally cropped, to discard the out-of-view area. The horizontal focal center is placed to be the center of the cropped image, resulting in a 1024$\times$1024 image. To fit our system's input size, the cropped image is then downsampled to a 256$\times$256 image.

The ground truth pose is adjusted to fit the unit used by the UnrealEgo. The UnrealEgo uses a unit length of centimeters for the ground truth pose, while the EgoCap dataset uses millimeters. 
EgoCap dataset consists of poses for 18 joints, including the head. But following the EgoCap paper, the head 3D pose is not estimated, resulting in a total of 17 joints, and 16 limbs. Note that the UnrealEgo dataset uses a different set of joints, which consists of 16 joints including the head.

In the real-world setting, the relative 3D transform of two cameras consists of the rotational component, thus the local pose is not the same for the two cameras. However, the 3D pose is provided only for the first camera's coordinates, so we use the same view-plane angle and 3D orientation computed in the first camera's coordinate system for both views.

\begin{figure}
    \includegraphics[scale=0.3]{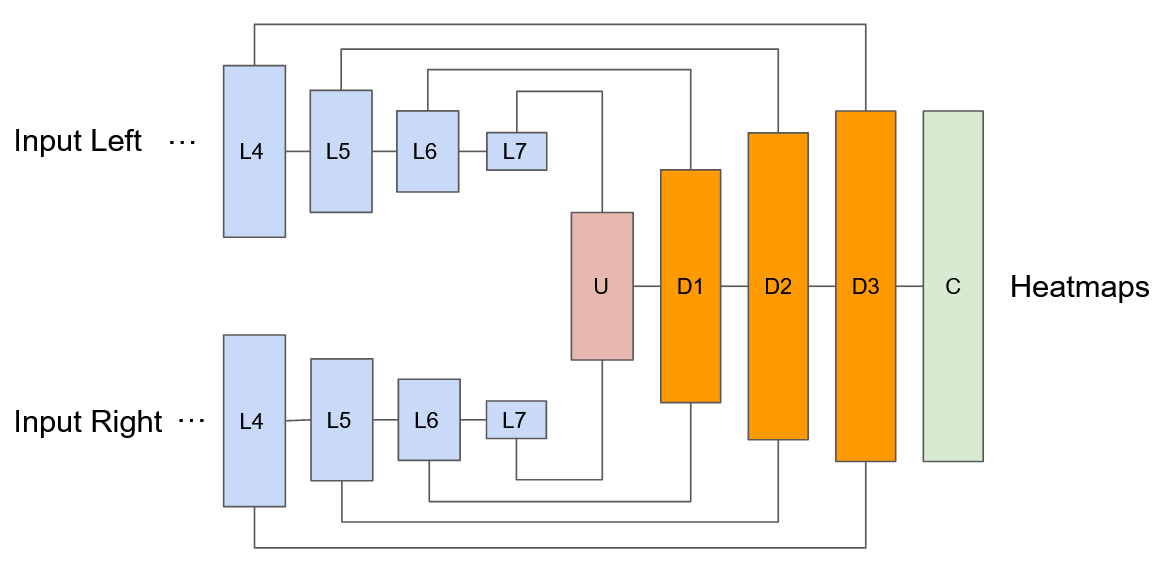}
    \caption{The architecture of the Optical Feature Extractor}
    \label{fig:optical_network}
\end{figure}

\begin{table}
 \caption{The hyperparameters for the decoder layers in the optical feature extractor.}
\small
\centering
\begin{tabular}{ |c||c|c|c| }
 \hline
 Layer & Dimension & Input Channels & Output Channels \\
 \hline
 D1 & 16$\times$16 & 3072 & 2048 \\
 D2 & 32$\times$32 & 2560 & 1024 \\
 D3 & 64$\times$64 & 1280 & 1024 \\
 \hline
\end{tabular}
 \label{table:optical_network}
\end{table}

\section{Network Details}
Our implementation is based on the open-sourced UnrealEgo's implementation. Each of the layers described as a convolutional, deconvolutional\cite{deconv2010}, and fully connected layer is followed by a batch normalization layer and a leaky ReLU with a negative slope of $0.2$.
\subsection{Optical Feature Extraction}
Two networks for the Joint Position Heatmaps and the Perspective Embedding Heatmap with the same architecture are trained for optical feature extraction. The ResNet-18~\cite{he2016residual} in the U-Nets~\cite{ronneberger2015convolutional} of the  optical feature extractors are initialized with pre-trained weights ImageNet1K\_1\cite{deng2009imagenet} available on PyTorch~\cite{NEURIPS2019_9015}.

Two ResNet are used in one U-Nets for each optical feature extractor to take stereo input images. In the torch's ResNet-18 implementation, the output of base layers of index 4, 5, 6, and 7 are concatenated to the U-Net's decoder parts, as shown in Fig.~\ref{fig:optical_network}. The output is processed by an 1 by 1 convolutional layer, to the same number of channels. In the decoder, the features from two ResNet base layer 7 are concatenated and upsampled after the process. The next layers take the concatenated processed output from two ResNets and the upsampled features from the previous layer. Each layer consists of one convolutional layer with the specified number of channels of kernel size 3, and an upsample layer following that. Table \ref{table:optical_network} describes the total input channels (the upsampled feature channels and the concatenated ResNet features) and output channels. Finally, one convolutional layer (C in Fig.~\ref{fig:optical_network}) takes the D3 layer's output and outputs heatmaps, using a kernel size of 1. 

\subsection{Heatmap Encoder}
The heatmap encoder consists of 3 convolutional layers and 3 fully-connected layers. The first convolutional takes all of the heatmaps. Each convolutional layer has 64, 128, and 256 channels of features, using a kernel size of 4, stride of 2, and padding of 1. The output of 256 channels of features is flattened and processed by the following fully connected layers.
Each fully connected layer has an output size of 2048, 512, and 20. The 20 is the size of the embedding vector used by the 3D Pose Decoder, and the Heatmap Reconstructor.

\subsection{3D Pose Decoder}
The 3D pose decoder consists of 3 fully connected layers. The first layer takes 20-dimensional embedding from the Heatmap Encoder and 14 by 3 estimated orientations from the Stereo Matcher, flattened and concatenated together as a vector. The first two layers output 32-dimensional embeddings, and the last layer outputs 16 by 3  estimated 3D pose.

\subsection{Stereo Matcher}
The stereo matcher module has a similar architecture to the combination of the Heatmap Encoder and the 3D Pose Decoder, with different input, intermediate embedding, and output sizes. The first difference is that it takes only 4 channels of heatmaps, the one set of Perspective Embedding Heatmaps. The output embedding size, from the Heatmap Encoder-like architecture is 10. The final decoder's output is a 3-dimensional vector, which corresponds to the estimated 3D orientation.

\subsection{Heatmap Reconstructor}
The Heatmap Reconstructor consists of 3 fully-connected layers and 3 deconvolutional layers. The fully connected layers take 20-dimensional embeddings and output 512, 2048, and one last 16384-dimensional vector that is reshaped to 256 channels of 8 by 8 features. The deconvolutional layer outputs 128, 64, and the total number of heatmaps channels in order. All of the deconvolutional layer uses a kernel size of 4, stride of 2, and padding of 1. The deconvolutional layer corresponds to PyTorch's torch.nn.ConvTranspose2d module.

\section{Limitations and Future Works}
There still remain several limitations. Occlusion in the egocentric pose estimation is still a challenging problem as many motions suffer from high occlusion, especially in the lower body. To deal with it, temporal optimization of the output poses is an important direction~\cite{wang2021estimating}. Additional inverse kinematics methods can be also useful for virtual character applications.

Secondly, the trained network can overfit the camera's distortion used in the training dataset. The 2D-to-3D lifting is an inherently ambiguous problem, without given camera parameters. Many egocentric methods focus on shared camera setups for 3D pose estimation. However, individual camera's distortion patterns may vary and the method can exhibit larger errors on cameras with different distortions. The Stereo Matcher network can introduce reliance on the binocular camera's configuration, as it attempts to estimate 3D pose from stereo correspondences. Recently, a 2D-to-3D lift-up model applicable for different camera optics and setups is suggested~\cite{omniego2022}. Such a generalizable framework that applies to various egocentric camera setups is a promising direction for future work.

Lastly, our real-world evaluation has several limitations in its variety, as a result of experimenting only on publicly available dataset. Evaluation in the real-world setting EgoCap dataset has a limited number of subjects and frames since we evaluated only the portion of the dataset with 3D pose annotations. It also lacks a variety of motion types, as it consists of activities while standing. Finally, due to the difficulty of egocentric dataset collection, the dataset is captured only in a lab environment with a green screen. The performance can further be experimented with more comprehensive real-world datasets.

\section{Additional Experiments}
\subsection{Impact of Using Joint Position Heatmap}
\label{subsubsec:impact-joint-heatmap}
We show the impact of using the Joint Position Heatmap when it is used together with the Perspective Embedding Heatmap. We experimented with our system with only one type of heatmap and both. For the Joint Position Heatmap-only experiment, the result in {\it Effectiveness of Perspective Embedding Heatmap} section is used. The result in Table \ref{table:joint_heatmap} reveals that using only Perspective Embedding Heatmap outperforms the method using only Joint Position Heatmaps by 4.7\% in MPJPE, and using both outperforms the latter by 8.8\%. Perspective Embedding Heatmap contains joint position information when it successfully estimates the 3D information. Its confidence is directly connected to its estimate of 3D angle. Thus, if the estimation of the 3D angle fails, the network may not output meaningful values for the positional estimate, even though the visual cue is available for the 2D position. In those cases, the traditional Joint Position Heatmap provides a fall-back option by focusing on extracting 2D information.

\begin{table}[h]
 \caption{Comparison of results on \textbf{UnrealEgo} dataset, using our system with Joint Position Heatmap (\textbf{JH}) only, Perspective Embedding Heatmap (\textbf{PH}) only, and both.}
\begin{tabular}{ |c||m{1.44cm}|m{1.44cm}|m{1.44cm}| }
 \hline
 \diagbox{Error}{Heatmaps} & JH & PH & JH + PH \\
 \hline
 MPJPE & 66.72 & 63.60 &\textbf{60.82} \\
 PA-MPJPE & 52.29 & 49.49 &\textbf{48.47} \\
 \hline
\end{tabular}
\hspace{0.15cm}
 \label{table:joint_heatmap}
\end{table}

\subsection{Per Joint Error Distribution}

\begin{figure}[h]
    \includegraphics[scale=0.5]{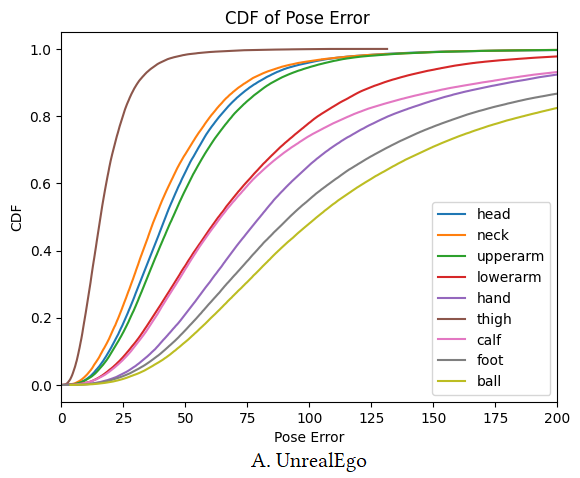}
    \includegraphics[scale=0.5]{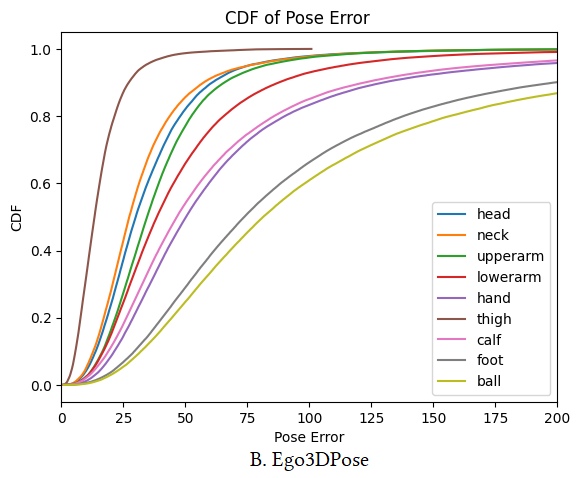}
    \caption{CDF of joint pose estimation error of the \textbf{UnrealEgo} (A) and \textbf{Ego3DPose} (B) in mm unit.}
    \label{fig:joint_dist_cdf}
\end{figure}

We plot the distribution of pose estimation error on the UnrealEgo dataset for two systems, UnrealEgo~\cite{hakada2022unrealego} and our Ego3DPose, in Figure.~\ref{fig:joint_dist_cdf}. In this plot, we combined the results of corresponding joints on the left and right as distinct samples for one category. "upperarm", "lowerarm", and "hand" corresponds to upper body joints, and "thigh", "calf", "foot", and "ball" are lower body joints. The distribution is visualized as a Cumulative Distribution Function (CDF).
As previous works~\cite{tome2019xr}\cite{zhao2021egoglass} suggest, lower body parts generally had larger estimation errors. In the experiment, however, the estimation of "thigh" appears to be accurate. This is due to the local pose's definition of the UnrealEgo dataset. The local pose is relative to the pelvis' position, and since the thighs are directly connected to the pelvis, it is easy to estimate its position.
Our method shows more improvement on the upper body, which has more visibility than the lower body, particularly noticeable when comparing "lowerarm", "hand", and "calf". It indicates that our method is more effective at extracting visible cues.

%% file: 99_figure_only.tex
\begin{figure*}[htp] 
    \includegraphics[scale=0.49]{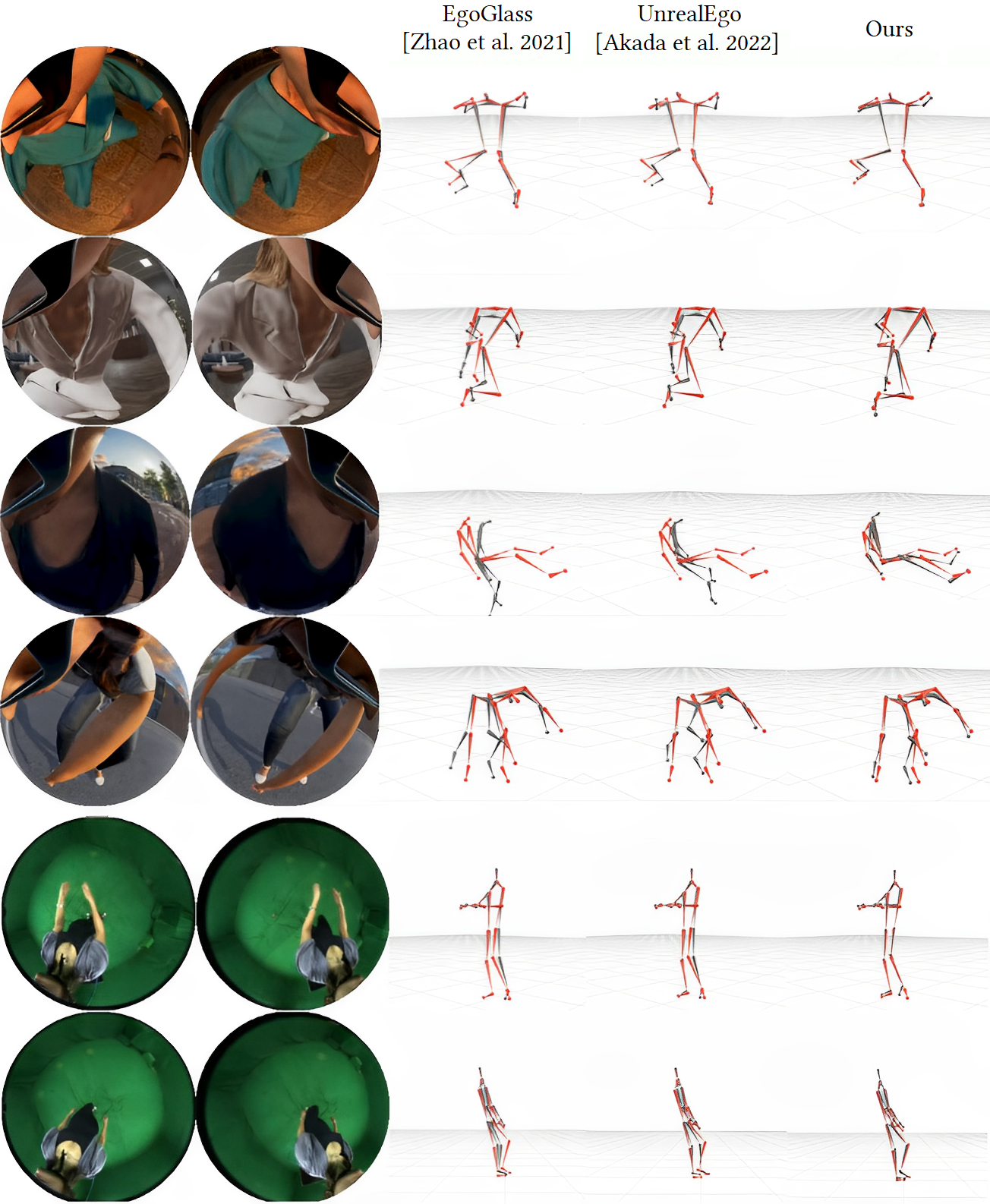}
    \caption{Qualitative evaluation of our method. The first four poses are from the synthetic UnrealEgo dataset, and the last two poses are from the real-world EgoCap dataset. The EgoCap validation 3D set has two sequences after our train/test set splits, so we show two samples from each. The red pose is the ground truth and the grey pose is the predicted pose from methods.}
    \label{fig:qualitative}
    \vspace{-0.3cm}
\end{figure*}